\documentclass[runningheads]{llncs}

\usepackage{accv}

\usepackage{accvabbrv}

\usepackage{graphicx}
\usepackage{booktabs}
\usepackage{multirow}

\usepackage[accsupp]{axessibility}  

\usepackage[pagebackref,breaklinks,colorlinks,citecolor=accvblue]{hyperref}

\usepackage{orcidlink}

\begin{document}

\title{Robust Latent Representation Tuning for Image-text Classification} 

\titlerunning{Robust Latent Representation Tuning}

\author{Hao Sun\inst{1}\orcidlink{0000-0001-8094-1991} \and
Yu Song\inst{2}}

\authorrunning{Hao Sun et al.}

\institute{College of Computer Science and Technology, Zhejiang University, Hangzhou, China 
\email{sunhaoxx@zju.edu.cn}\\
\and College of Information Science and Engineering, Ritsumeikan University, Osaka, Japan\\
\email{yusong@fc.ritsumei.ac.jp}}

\maketitle

\begin{abstract}
Large models have demonstrated exceptional generalization capabilities in computer vision and natural language processing. Recent efforts have focused on enhancing these models with multimodal processing abilities. However, addressing the challenges posed by scenarios where one modality is absent remains a significant hurdle. In response to this issue, we propose a robust latent representation tuning method for large models. Specifically, our approach introduces a modality latent translation module to maximize the correlation between modalities, resulting in a robust representation. Following this, a newly designed fusion module is employed to facilitate information interaction between the modalities. Within this framework, common semantics are refined during training, and robust performance is achieved even in the absence of one modality. Importantly, our method maintains the frozen state of the image and text foundation models to preserve their capabilities acquired through large-scale pretraining. We conduct experiments on several public datasets, and the results underscore the effectiveness of our proposed method.
\keywords{Image-text classification \and large models \and robust learning \and representation learning}
\end{abstract}

\section{Introduction}
\label{sec:intro}
In recent times, large models have garnered substantial attention due to their remarkable generalization capabilities across numerous downstream tasks. Given that most large models are pretrained on unimodal datasets (e.g., LLaMA~\cite{touvron2023llama}, OPT~\cite{zhang2022opt}), researchers have sought to augment these models with multimodal processing capabilities. Notably, approaches like LISA~\cite{lai2023lisa} have proposed extracting multimodal features using various large models, employing these features for tasks such as image segmentation. PixelLM~\cite{ren2023pixellm} has introduced a tuning framework wherein visual embeddings are prefixed to textual tokens, jointly processed by large language models (LLM). Despite the numerous endeavors to imbue large models with the capacity to process multimodal signals (e.g., images and texts), there has been limited attention to robust representation learning, and performance in modality-absence scenarios remains relatively unexplored. Real-world applications frequently encounter scenarios where certain modalities are absent or noised, rendering current methods challenging to apply.

To address this challenge, we introduce a novel strategy for robust multimodal representation tuning in this paper. Our approach leverages two pretrained large models dedicated to image and text processing. At each corresponding layer of the paired image-text models, we incorporate a Modality Latent Translation (MoLT) module. Within this module, image and text embeddings are projected onto a shared latent space for cross-modality interaction. This shared space acts as a bridge connecting the image and text domains. Subsequently, we employ factorized bilinear pooling (FBP)\cite{yu2017multi} to obtain the robust representation, which has been proven effective in multimodal learning\cite{sun2023modality}. After feature extraction, a cross-attention mechanism is employed to capture the relationship between the robust representation and the associated modality embeddings for making predictions.

At the heart of our method, MoLT comprises two cross-attention modules, individually tailored for the image and text domains. Following the cross-attention step, we apply a Canonical Correlation Analysis (CCA) loss~\cite{andrew2013deep} to facilitate the learning of a robust representation between the two modalities. Consequently, in scenarios where one modality is absent, a straightforward translation from the available modality or the utilization of the learned robust representation becomes feasible for downstream tasks. Throughout our training process, the parameters from pretrained models remain frozen, allowing only the newly introduced modules to be tunable. This approach enables the model to progressively acquire and refine robust representations.

In summary, our contributions can be outlined as follows:
\begin{itemize}
    \item We propose a novel strategy for robust representation tuning in large models. Our method facilitates the learning of a robust representation in a shared latent space, establishing a bridge between image and text embeddings.
    \item Introducing the Modality Latent Translation (MoLT) module in our approach, we present a sophisticated cross-attention module that brings text and image embeddings closer together.
    \item Our model achieves state-of-the-art performance on evaluated image-text classification datasets. Furthermore, our experiments demonstrate the model's remarkable robustness in scenarios involving modality absence.
\end{itemize}

\section{Related Works}
\label{sec:related}
In this section, we introduce some recent works related to our method, including large models and corresponding tuning method.

\subsection{Large Vision and Language Models}
The advent of large models has dominated discussions in deep learning, particularly within the realms of computer vision and natural language processing. Noteworthy language models include GPT-3~\cite{floridi2020gpt}, LLaMA~\cite{touvron2023llama}, and OPT~\cite{zhang2022opt}, which, pretrained on extensive corpora, exhibit formidable capabilities in comprehending and generating long-context information. In the domain of computer vision, SAM~\cite{kirillov2023segment} stands as the current state-of-the-art foundation model for visual understanding. However, the scarcity of open-source large models trained on multimodal corpora, such as CLIP~\cite{radford2021learning}, poses challenges for processing multimodal data using large models.

\subsection{Multimodal Large Model Tuning}
Recent years have witnessed a surge of interest in the tuning of large models. While most tuning strategies are devised for unimodal processing, some researchers have endeavored to integrate multimodal information into large models through multimodal tuning. For instance, Flamingo~\cite{alayrac2022flamingo} proposes fusing multimodal signals with gated cross-attention into a frozen image encoder, showcasing the potential of large models for multimodal processing. BLIP~\cite{li2022blip} aligns multimodal embeddings through multitask learning, and BLIP-2~\cite{li2023blip}, subsequently proposed with a Q-Former, finds widespread application in recent works. PaLM-E~\cite{driess2023palm} introduces sending visual tokens as input to pretrained language models, demonstrating impressive performance. In FROMAGe~\cite{koh2023grounding}, researchers explore grounding texts and images to each other to attain multimodal understanding capabilities. Our proposed method also focuses on tuning large models but places a distinct emphasis on robust representation learning.

\begin{figure*}[ht]
\begin{minipage}[b]{1.0\linewidth}
  \centering
  \includegraphics[width=1.0\textwidth]{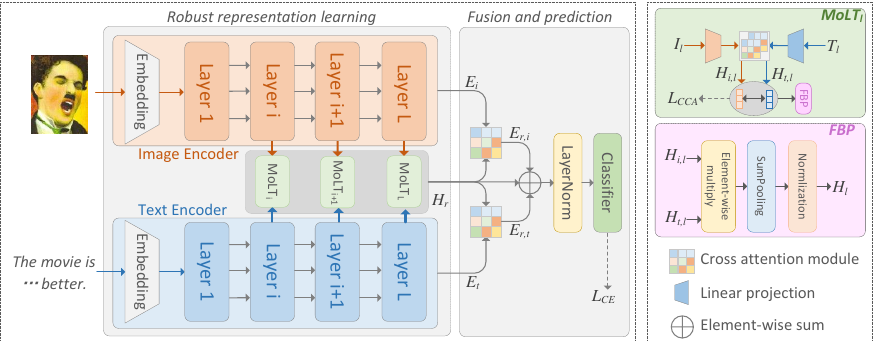}
\end{minipage}
\caption{The overview of our proposed method. The image and text are first processed by separate encoders for robust representation learning. During this process, the robust representation $H_r$ is obtained by MolT module and FBP. After that we fuse the modality features and robust embedding for the final predictions.}
\label{fig:overview}
\end{figure*}

\section{Method}
\label{sec:method}
The pipeline of our proposed approach is illustrated in Figure~\ref{fig:overview}. Our method comprises two main modules for image-text classification. Given an image-text pair ($I$, $T$), we initially dispatch them to their corresponding frozen foundation models for feature extraction. In this stage, a modality latent translation module is introduced to facilitate robust representation learning. Subsequently, the obtained robust representation, in conjunction with text and image embeddings, is integrated for final predictions through our newly designed structure.

\subsection{Modality Latent Translation}
When the image and text are processed by respective large models, a modality latent translation module(MolT) is introduced to learn the robust representation. For each pair of image-text foundation model layers $l$, the corresponding representations are $I_{l}\in R^{N_{i}\times d_{i}}$ and $T_{l}\in R^{N_{t}\times d_{t}}$, where $N$ and $d$ are respective token numbers and dimensions. Then two linear projections are employed to map the embeddings into the same space:
\begin{equation}
    \begin{aligned}
        I_{l}^{'} = W_{i}\cdot I_{l} + b_{i} \in R^{N_{i}\times d_{c}},\\
        T_{l}^{'} = W_{t}\cdot T_{l} + b_{t} \in R^{N_{t}\times d_{c}},\\
    \end{aligned}
\end{equation}
where $W_{i}\in R^{d_{i}\times d_{c}}$, $W_{t}\in R^{d_{t}\times d_{c}}$, $b_{i}\in R^{d_{c}}$, and $b_{t}\in R^{d_{c}}$ are four learnable parameters, and $d_{c}$ is the dimension of common space. Followingly, two cross-attention module is employed to perform the modality interaction between two embeddings:
\begin{equation}
    \begin{aligned}
        H_{i,l} = SoftMax(\frac{Q_{i}K_{t}^{T}}{\sqrt{d_{c}}})V_{t},\\
        H_{t,l} = SoftMax(\frac{Q_{t}K_{i}^{T}}{\sqrt{d_{c}}})V_{i},\\
    \end{aligned}
\end{equation}
where $Q_{i}$, $K_{i}$, and $V_{i}$ are transformed modality embeddings from image embedding $I_{l}^{'}$. The same is true for $Q_{t}$, $K_{t}$, and $V_{t}$. Through cross-attention, the modality embeddings gain access to information from each other, yielding a more comprehensive set of common semantics. The resulting normalized embeddings, augmented with interacted residuals, are then regarded as the representation of each modality in the common space:
\begin{equation}
    \begin{aligned}
        H_{i,l}^{'} = Norm(H_{i,l} + I_{l}^{'}),\\
        H_{t,l}^{'} = Norm(H_{t,l} + T_{l}^{'}).\\
    \end{aligned}
\end{equation}
Finally, we try to maximize the canonical correlation between $H_{i,l}^{'}$ and $H_{t,l}^{'}$ via DCCA~\cite{andrew2013deep}, so as to bring them closer to each other in the common space. Specifically, let $R_{11}$, $R_{22}$ be variances of $H_{i,l}^{'}$ and $H_{t,l}^{'}$, the covariance between $H_{i,l}^{'}$ and $H_{t,l}^{'}$ as $R_{12}$. The canonical-correlation analysis(CCA) loss can be defined by:
\begin{equation}
    \begin{aligned}
        L_{CCA} = -trace(F^{T}F)^{0.5},
    \end{aligned}
\end{equation}
where $F=R_{11}^{-0.5}R_{12}R_{22}^{-0.5}$. Throughout the tuning process, the representations become more robust as the canonical correlation increases. Consequently, we can effectively translate representations from one modality to another, thereby empowering the model with the ability to infer in scenarios where one modality is missing. Finally, we obtain the robust representation of layer $l$ by factorized bilinear pooling(FBP):
\begin{equation}
    \begin{aligned}
        H_{l}^{'} &= SumPooling(H_{i,l}^{'} \odot H_{t,l}^{'}, s),\\
        H_{l} &= \frac{H_{l}^{'}}{||H_{l}^{'}||_{2}},
    \end{aligned}
\end{equation}
where $\odot$ means the element-wise multiply and $s$ is the stride of sum pooling. The robust representation $H_{l}$ in each layer are utilized in the fusion stage for robust predictions.

\subsection{Fusion and Training Target}
After the frozen foundation model, the robust representations ($H_{l}$) and extracted modality embeddings($E_{i}$, and $E_{t}$) are sent to our new designed fusion. The detailed structure is shown in Figure~\ref{fig:overview}. We first introduce a learnable vector $M\in R^{L_{s}}$ to average-pool the robust representations from each selected image-text layers:
\begin{equation}
    \begin{aligned}
        H_{r} &= Avg(M \cdot [H_{1}, H_{2}, ..., H_{l}, ..., H_{L_{s}}]),\\
    \end{aligned}
\end{equation}
where $L_{s}$ is the number of selected layers for robust representation learning. Then we perform the information exchange between $H_{r}$, $E_{i}$, and $E_{t}$:
\begin{equation}
    \begin{aligned}
        E_{r,i} = SoftMax(\frac{Q_{r}K_{i}^{T}}{\sqrt{d_{c}}})V_{i},\\
        E_{r,t} = SoftMax(\frac{Q_{r}K_{t}^{T}}{\sqrt{d_{c}}})V_{t},\\
    \end{aligned}
\end{equation}
where $Q_{r}$ is projected by $H_{r}$, $K_{i}$/$V_{i}$ are projected from $E_{i}$, and $K_{t}$/$V_{t}$ are projected from $E_{t}$. To avoid the instability caused by the absence of a modality, we apply layer normalization before making the final predictions:
\begin{equation}
    \begin{aligned}
        \hat{y} = Classifier(\frac{1}{2}Norm(E_{r,i} + E_{r,t})).
    \end{aligned}
\end{equation}

In our method, we employ two training targets: the CCA loss and the task loss. The final loss function can be represented by:
\begin{equation}
    \begin{aligned}
        L = \alpha L_{CCA} + \beta L_{CE},
    \end{aligned}
\end{equation}
where $L_{CE}$ is the cross entropy loss, $\alpha$ and $\beta$ are two hyper-parameters to balance the numerical scales of different losses.

\section{Experiments and Analysis}
In this section, we introduce our experimental settings, evaluated benchmarks and respective results.

\subsection{Benchmark Datasets}
To evaluate the effectiveness of our proposed method, we conduct the experiments on three public datasets: MM-IMDB~\cite{arevalo2017gated}, UPMC-Food101~\cite{wang2015recipe}, and  SNLI-VE~\cite{xie2018visual}. Among the three datasets, \textbf{MM-IMDB} dataset is to classify the movie into one or more of the 23 genres with the poster image and textual outlines. This dataset contains contains 15510 training samples, 2599 validation samples and 7779 samples for test. \textbf{UPMC-Food101} is a popular image-text classification dataset, which aims to categorize food images with recipe descriptions into 101 categories. There are 67971 training samples and 22715 test samples. \textbf{SNLI-VE} is a visual-entailment understanding dataset, in which each sample includes an image premise and a text hypothesis. The labels are annotated by the semantic relationship(entailment, neutral, or contradiction) between them. The datasets contains 529527 samples for training, 17585 for validation, and 17901 for test.

\subsection{Experimental Settings}
In our experiments, we employ the pretrained LLaMA as the text foundation model and the image encoder of CLIP-L/224 as visual foundation model. Inherited from the pretrained models, $d_{i}$ and $d_{t}$ are set to 4096 and 768, respectively. The dimension of common space $d_{c}$ is set to 1024. We infuse the MoLT module in the last 4 layers of image and text models, meaning that $L_{s}$ is 4. $s$ is set to 4 in the FBP operations. In the loss function, we set $\alpha$ to 0.1 and $\beta$ to 0.9. To reduce the memory consumption, we train our model with mixed-precision. The Adam optimizer is employed in our method and the learning rate is set to 0.0004. Our approach is implemented with PyTorch framework and the experiments are conducted on two NVIDIA RTX 3090Ti GPUs.

\begin{table*}[ht]
\centering
\caption{The quantitative results of our method on three benchmark datasets. \textit{w/ LM} indicates whether the large models are utilized in the approach. * indicates that the original paper did not provide corresponding results, and we re-implement the method based on their open-source codes.}
\label{tab:result}
\begin{tabular}{ccccc}
\toprule
\multirow{2}{*}{Method} & \multirow{2}{*}{w/ LM} & MM-IMDB & UPMC-Food101 & SNLI-VE \\
 &  & F1-micro/macro(\%) & Acc(\%) & Acc(\%) \\
\midrule
HUSE~\cite{narayana2019huse} &  & 62.1*/54.2* & 92.30 & 69.90* \\
VisualBERT~\cite{jia2022visual} &  & 63.4*/55.9* & 92.30 & 75.06 \\
\midrule
MMBT~\cite{kiela2019supervised} & $\checkmark$  & 66.8 / 61.8 & 92.10 & 74.69 \\
MaPLe~\cite{khattak2023maple} & $\checkmark$ & 60.9 / 51.2 & 90.80 & 71.52 \\
BlindPrompt~\cite{liang-etal-2022-modular} & $\checkmark$ & 56.5 / 50.2 & 84.56 & 65.54 \\
PMF~\cite{li2023efficient} & $\checkmark$ & 64.5 / 58.8 & 91.51 & 71.92 \\
Ours & $\checkmark$ & \textbf{67.0}/\textbf{61.9} & \textbf{92.44} & \textbf{75.10} \\
\bottomrule
\end{tabular}
\end{table*}

\subsection{Quantitative Results}
The results of our method on the evaluated datasets are presented in Table~\ref{tab:result}. As evident from the results, we achieve state-of-the-art performance on each benchmark, demonstrating the effectiveness of our approach. Among the methods we compare with, HUSE~\cite{narayana2019huse} and VisualBert~\cite{jia2022visual} do not utilize large models, which highlights the advantage of incorporating advanced model architectures. In contrast, other methods, such as MaPLe~\cite{khattak2023maple}, MMBT~\cite{kiela2019supervised}, and PMF~\cite{li2023efficient}, are based on large models. Most large model-based methods aim to facilitate information exchange through fine-tuning strategies. Specifically, MaPLe~\cite{khattak2023maple} proposed a tuning module that maps linguistic embeddings to the visual branch, enhancing cross-modal interactions. Meanwhile, PMF~\cite{li2023efficient} aims to disentangle vanilla prompts into different types for tuning large models, optimizing the adaptability and specificity of the prompts to various tasks. The substantial performance gap between our method and previous methods illustrates the effectiveness of our proposed technique, showcasing its potential to significantly advance the field. Our approach not only leverages the strengths of large model architectures but also introduces innovative fine-tuning mechanisms that enhance model performance across diverse datasets.

\begin{table}[ht]
\centering
\caption{The ablation study on SNLI-VE and UPMC-Food101 dataset. \textit{C.A.} means the cross-attention in MoLT. \textit{Fusion} indicates the fusion strategy in our method.}
\label{tab:ablation}
\begin{tabular}{ccccc|cc}
\toprule
\multicolumn{5}{c|}{Ablation} & SNLI-VE & UPMC-Food101 \\
C.A. & $L_{CCA}$ & $M$ & Fusion & FBP & Acc(\%) & Acc(\%) \\
\midrule
 & $\checkmark$ & $\checkmark$ & $\checkmark$ & $\checkmark$ & 70.6 & 89.98 \\
$\checkmark$ &  & $\checkmark$ & $\checkmark$ & $\checkmark$ & 72.5 & 91.03 \\
$\checkmark$ & $\checkmark$ &  & $\checkmark$ & $\checkmark$& 74.0 & 91.35 \\
$\checkmark$ & $\checkmark$ & $\checkmark$ &  & $\checkmark$& 72.9 & 90.67 \\
$\checkmark$ & $\checkmark$ & $\checkmark$ & $\checkmark$ & & 74.21 & 92.65 \\
$\checkmark$ & $\checkmark$ & $\checkmark$ & $\checkmark$ & $\checkmark$ & \textbf{75.10} & \textbf{92.44}\\
\bottomrule
\end{tabular}
\end{table}

\subsection{Ablation Study}
To further investigate the effectiveness of each component in our method, we conducted an ablation study using SNLI-VE and UPMC-Food101 datasets. The results are presented in Table~\ref{tab:ablation}. When removing $L_{CCA}$, the performance drops dramatically, highlighting the crucial role of this training target in aligning the multimodal embeddings. This significant decrease underscores how central $L_{CCA}$ is to the model’s ability to capture and integrate cross-modal correlations. The results also demonstrate that both the cross-attention module and the learnable vector $M$ contribute positively to the final outcomes. The cross-attention module facilitates nuanced interactions between modalities, while the learnable vector $M$ adapts dynamically to enhance representation quality. When the fusion module is omitted, meaning only the robust representations are employed for predictions, there is a noticeable decrease in performance. This decline underscores the importance of the fusion module in synthesizing information from multiple modalities effectively. Correspondingly, the results also illustrate the effectiveness of the FBP module, which aims to integrate multimodal embeddings into a unified, robust representation. The FBP module's role in enhancing the coherence and informativeness of the combined features is clearly evidenced by the performance drop observed when it is removed.

The performance gap observed between each ablative model and our final model underscores the effectiveness of our proposed method. This gap highlights how each component, from $L_{CCA}$ to the fusion and FBP modules, plays a pivotal role in achieving state-of-the-art results. Our comprehensive ablation study confirms that the synergistic integration of these components is essential for maximizing the performance of our method on the SNLI-VE dataset.

\begin{figure}[ht]
\centering
\begin{minipage}[b]{1.0\linewidth}
  \centering
  \includegraphics[width=1.0\textwidth]{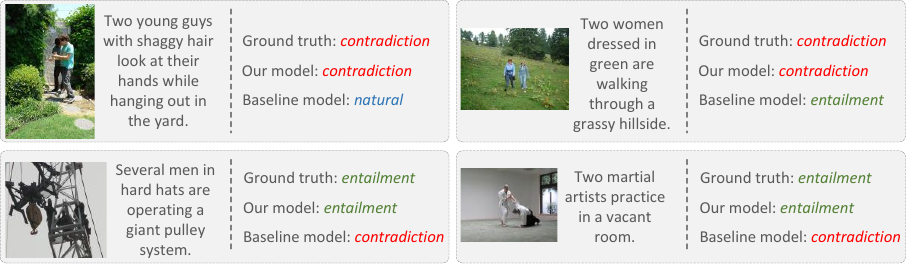}
\end{minipage}
\caption{The visualization of some cases for our propose method and the baseline model. The examples are from the SNLI-VE dataset.}
\label{fig:visual}
\end{figure}

\begin{table*}[ht]
\centering
\caption{The inference performance of our method under modality-missing and noisy scenarios. In noised-modality inference, a certain proportion of Gaussian noise is added to the original input. The baseline model indicates neither the MolT module nor robust representations were utilized for final predictions.}
\label{tab:robust}
\begin{tabular}{c|ccc}
\toprule
\multirow{2}{*}{Models} & MM-IMDB & UPMC-Food101 & SNLI-VE \\
 & F1-micro/macro(\%) & Acc(\%) & Acc(\%) \\
\midrule
\multicolumn{4}{c}{Modality-absence Inference} \\
\midrule
Baseline(text-absence) & 57.1 / 47.9 & 82.1 & 67.4  \\
Baseline(image-absence) & 59.0 / 51.2 & 85.2 & 69.3  \\
Ours(text-absense) & 62.4 / 56.7 & 88.9 & 73.2  \\
Ours(image-absense) & 63.1 / 57.0 & 89.2 & 71.0 \\
\midrule
\multicolumn{4}{c}{Noised-modality Inference} \\
\midrule
Baseline(1\% noise)  & 59.2 / 50.5 & 82.2 & 65.5 \\
Baseline(5\% noise)  & 56.6 / 49.7 & 80.9 & 61.0  \\
Baseline(10\% noise) & 48.1 / 39.9 & 77.1 & 55.8 \\
Baseline(20\% noise) & 36.9 / 27.5 & 52.0 & 39.3  \\
Ours(1\% noise)      & 63.0 / 56.3 & 87.1 & 68.2  \\
Ours(5\% noise)      & 59.9 / 54.2 & 86.0 & 62.2 \\
Ours(10\% noise)     & 51.1 / 43.3 & 79.8 & 56.1 \\
Ours(20\% noise)     & 40.0 / 31.2 & 53.3 & 39.4 \\
\bottomrule
\end{tabular}
\end{table*}

\subsection{Robust Inference Analysis}
As our focus lies in enhancing model robustness in modality-missing inference scenarios, we conducted corresponding experiments to evaluate our approach under challenging conditions. These experiments were performed under two distinct scenarios: one involving missing modalities and the other involving noised modality signals(we manually add a certain proportion of Gaussian noise to the input). In the baseline model, neither the MolT module nor robust representations were utilized for predictions. The corresponding results are shown in Table~\ref{tab:ablation}.

The results indicate that each modality has a varying impact on different tasks. For instance, in movie classification, the textual modality proves to be dominant, whereas the image modality becomes more critical for tasks like visual-entailment understanding. Despite these variations, our method consistently outperforms the baseline model, demonstrating the effectiveness of our proposed approach across different tasks. Moreover, even when the model is tested with noised multimodal features, our proposed method still yields relatively strong performance, whereas the baseline model's performance drops dramatically. This resilience to noisy inputs highlights the robustness of our approach. We further substantiate our findings through visual inspections of several cases, as shown in Figure~\ref{fig:visual}. These visual inspections reveal that with the incorporation of the MolT module and robust representation learning, our model can still make accurate predictions, whereas the baseline model often fails.

\section{Conclusion}
In this paper, we propose a robust representation learning strategy tailored for large models. Our approach incorporates a modality latent translation module capable of translating one modality embedding to another, with the factorized bilinear pooling used to generate robust representations. Additionally, we introduce a novel fusion schema for robust representation and modality embeddings. Experiments conducted on three datasets clearly illustrate the effectiveness of our proposed method. Further analysis demonstrates the robustness of our method in modality-missing and noisy scenarios. In the future, we plan to conduct further research in robust representation learning to enhance our ability to handle modality-absence scenarios more effectively.

%
%
\bibliographystyle{splncs04}
\bibliography{main}
\end{document}